\documentclass[conference]{IEEEtran}
\IEEEoverridecommandlockouts
\usepackage{cite}
\usepackage{fancyhdr}
\usepackage{amsmath,amssymb,amsfonts}
\usepackage{algorithmic}
\usepackage{graphicx}
\usepackage{subfig}
\usepackage{textcomp}
\usepackage{xcolor}
\def\BibTeX{{\rm B\kern-.05em{\sc i\kern-.025em b}\kern-.08em
    T\kern-.1667em\lower.7ex\hbox{E}\kern-.125emX}}

\begin{document}

\title{Chromosome Segmentation Analysis Using Image Processing Techniques and Autoencoders}

\author{Amritha S Pallavoor\textsuperscript{1}, Prajwal A\textsuperscript{1}, Sundareshan TS\textsuperscript{2}, Sreekanth K Pallavoor\textsuperscript{3}\\
{\footnotesize \textsuperscript{1}PES University, Bangalore, India  \footnotesize \textsuperscript{2}Dr. Rao’s Genetics Laboratory and Research Center, Bangalore, India  \footnotesize \textsuperscript{3}CARAIO Technologies, Bangalore, India}}

\maketitle
\begin{abstract}
Chromosome analysis and identification from metaphase images is a critical part of cytogenetics based medical diagnosis. It is mainly used for identifying constitutional, prenatal and acquired abnormalities in the diagnosis of genetic diseases and disorders. The process of identification of chromosomes from metaphase is a tedious one and requires trained personnel and several hours to perform. Challenge exists especially in handling touching, overlapping and clustered chromosomes in metaphase images, which if not segmented properly would result in wrong classification. We propose a method to automate the process of detection and segmentation of chromosomes from a given metaphase image, and in using them to classify through a Deep CNN architecture to know the chromosome type. We have used two methods to handle the separation of overlapping chromosomes found in metaphases - one method involving watershed algorithm followed by autoencoders and the other a method purely based on watershed algorithm. These methods involve a combination of automation and very minimal manual effort to perform the segmentation, which produces the output. The manual effort ensures that human intuition is taken into consideration, especially in handling touching, overlapping and cluster chromosomes. Upon segmentation, individual chromosome images are then classified into their respective classes with 95.75\% accuracy using a Deep CNN model. Further, we impart a distribution strategy to classify these chromosomes from the given output (which typically could consist of 46 individual images in a normal scenario for human beings) into its individual classes with an accuracy of 98\%. Our study helps conclude that pure manual effort involved in chromosome segmentation can be automated to a very good level through image processing techniques to produce reliable and satisfying results.

\end{abstract}
 
\let\thefootnote\relax\footnote
{Corresponding Author: Sreekanth K. Pallavoor, Founder CARAIO Technologies
(email: sreekanth@caraiotech.com)
}

\section{Introduction}
Human Chromosome Analysis involves the identification of 23 pairs of Chromosomes in human cells, out of which the first 22 pairs are called Autosomes and the 23rd pair is the Sex Chromosome. This process is called Karyotyping, and it is typically performed by experts in the field of Cytogenetics.

Its potential is immense in the early detection/diagnosis of diseases including Constitutional abnormalities, Prenatal and Acquired abnormalities.
However, the whole process is very manual, labour intensive and hence prone to human errors and fatigue, which could effectively result in delays in report generation and inaccurate/faulty reports.

The process of Karyotyping includes multiple stages, and the whole exercise involves a lot of manual effort to go through the images of chromosomes, look for abnormalities if any and confirm those and prepare the final report. The adoption of Artificial Intelligence and Image Processing is pertinent to this problem domain, where information exists in terms of digital images.

The focus of this paper is to develop an automated image segmentation process that identifies and segments the individual chromosomes (46 in number for normal cases) from the metaphase image obtained through a microscope, and further classify them with a deep learning model. This process when implemented through a software is intended to help freshers and students in cytogenetics to learn to use their intuition and domain knowledge to separate overlapping chromosomes in an easier and effective way through combining human and machine intelligence, thereby making their training process much simplified. Additionally, it can function just as an aid in diagnosis as well.

Routine chromosome analysis in cytogenetics requires culturing of samples for a certain period depending upon the sample type, followed by metaphase slide preparation and image capture of metaphase using microscope. This process involves one of the various staining methods, out of which we are considering here the case involving Q-banding staining method and the G-banding staining method separately. The Q-banding staining method is a fluorescent staining method, which uses quinacrine, is used to identify individual chromosomes and their structural anomalies, whereas G-banding is done by using Giemsa or Leishman stains to produce thin, alternating bands along the length of the entire chromosome that create unique patterns through which identification is done. The characteristic banding pattern can be used to identify each chromosome accurately. These images are then loaded on to a software and manually analysed to segment them using human expertise and through imparting visual cognitive effort. We propose to automate this part through implementing Image Processing techniques to segment chromosomes more efficiently, especially for early stage users in the field. After segmentation of the chromosome images through this approach, we propose to use these images with a Deep CNN architecture to train and achieve the automated classification of these chromosome images to a high accuracy level to produce the Karyogram output.

Our contribution is 4 fold:
\begin{enumerate}
    \item Identification and separation of individual chromosomes.
    \item Detecting and separating touching, overlapping, and chromosome clusters.
    \item Predicting the chromosome classes for all individual chromosome images also handling abnormality detection.
    \item A distribution algorithm that increases overall accuracy.

\end{enumerate}

\section{Related Work}
The analysis of a Karyotype has a wide range of uses in the cytogenetics field, its most important and common usage lies in prenatal diagnosis and genetic disease detection. 
The current industry method involves manual segmentation and classification by a trained person to classify and detect anomalies. 

Our methods and research for chromosome image segmentation is performed on G-Band and Q-Band metaphase images. Classification of chromosomes is constrained to Q-Band metaphase images as that’s where we achieved better accuracy. We are continuing our research on G-Band image classification to obtain the best possible accuracy.

%Statistical Karyotype Analysis Using CNN and Geometric Optimization
[1] proposed the use of multiple input convolutional neural networks (CNN) and geometric optimization, called mCNN\_GO for classification. They used Mask R-CNN for the segmentation of individual chromosomes from the metaphase images and classified the sub-images using the mCNN\_GO. They also performed chromosome straightening with a medial axis locating algorithm, and achieved around 95.644\% accuracy for segmentation.

%Automated System for Chromosome Karyotyping to Recognize the Most Common Numerical Abnormalities Using Deep Learning
[2] proposed a segmentation method where individual chromosome detection was done using YOLOv2 CNN followed by some chromosome post-processing. This step achieved 0.84 mean IoU. They used VGG19 for further processing and classification and obtained an accuracy of 94.11\% on the BioImLab Q-Band image dataset. They worked with metaphase images containing non-overlapped chromosomes.

%Overlapping Chromosome Segmentation Using UNET: Convolutional Networks with Test Time Augmentation
[3] By adding a number of layers onto the U-Net architecture, they performed overlapping chromosome semantic segmentation by implementing TTA and reached 99\% accuracy.

[4] detected overlaps by performing thinning of the image using a Morphological operation. They found the cut points of the intersection by implementing an algorithm with a predefined 7x7 mask. 

[5] used features extraction methods, and extracted medial axis, polarisation and length of individual chromosomes, to then classify them using an MLP network, they also used a multi-stage decision tree to find the polarisation of the chromosome. They achieved an accuracy of 95.6\%

[6] proposed using extensive features that were extracted from the chromosome images, and improved estimation of the medial axis. Feature re-scaling and normalising techniques take full advantage of the results of the polarisation step, reducing the intra-class and increasing the inter-class variances. They also use a rule-based approach that works on features to identify the polarisation. An MLP is used for classification and the accuracy obtained is 94\%

Both [5] and [6] employ a distribution strategy that takes into account the number of chromosomes constraint and helps them classify chromosomes from a karyotype with better accuracies.

[7] proposed an ensemble of 3 Deep CNN models to further increase the classification accuracies. Their method achieves a classification accuracy of 97\% with the ensemble of VGG19, ResNet50 and MobileNetv2

[8] proposed a multistage architecture that includes a network that upscales the image to higher resolution, using super-resolution layers to upscale. They further use an Xception or a ResNet50 to classify the scaled images. 
The highest accuracy was achieved by using Xception and it is 93\%

[9] proposed using a super-resolution net and self-attention negative feedback network and combining it with Deep CNNs to obtain a classification method (SRAS-net). They tackled the class imbalance of X and Y chromosomes by using SMOTE to generate additional Y samples. They achieved an accuracy of 97.5\%

[16] focused on crowd sourcing, and enabling large segmented dataset which can be fed to DNN models for automatic classification. However, this is dependent on human effort a lot, while some automation is there. Human in the loop is a key consideration factor here.

[17] The distinction between partially overlapping chromosomes was done by using a neural network based image segmentation method. This was applied on a synthetic dataset by using U-Net for segmentation. The results achieved IoU scores of 94.7\% for the overlapping region and 88-94\% for the non-overlapping chromosome regions.

\section{Dataset}
We used two sets of chromosome image dataset for the research and development of this project. 
The first dataset includes Q Band chromosome images which are obtained by staining chromosomes with Quinacrine, a fluorescent dye in the laboratory. It consists of 230 images of individual chromosomes for each type, across all 23 types of chromosome classes (including X and Y as class 23). The dataset also has 117 metaphase images from which these individual chromosome images are taken. These images include a wide range of chromosome orientations which are straight, bent and so on and also include touching, overlapping and clusters of chromosomes in the metaphase images. The individual chromosome images are all in the right polarity with the p-arm placed above the q-arm, and was used for the training of our deep learning models to predict the chromosome classes. This is a publicly available dataset found online from the BioImLab - Laboratory of Biomedical Imaging. Refer to [10] for more information about this dataset.\\
The second dataset we used contains G-Band metaphase images which are prepared by staining chromosomes with the Giemsa stain. Here, the banding patterns are more pronounced and can be seen more clearly when compared to Q-Band stained images. We obtained this dataset from Dr. Rao's Genetics Laboratory and Research Center, Bangalore, India. The metaphase images also include a wide range of chromosomes that are touching, overlapping or clustered.

\section{Methodology}
The metaphase image comprises 46 chromosomes for a normal human cell or it maybe 45 or 47 chromosomes typically in cells involving numerical abnormalities. These chromosomes are scattered in the image and they occur in ways that require complex segmentation methods. We have categorized them into 4 main categories-
\begin{enumerate}
    \item Individual isolated chromosomes
    \item Touching chromosomes 
    \item Overlapping chromosomes
    \item Clusters of chromosomes
\end{enumerate}

\subsection{Segmentation - Initial Stage}
The G-Band metaphase images (refer Fig. 1) occur with pixel intensities varying from 0 to 255. It generally includes chromosomes in gray lying on a white background. We first perform object detection to locate the chromosomes on the image. Noise reduction is done to the image by applying the medianBlur filter over the image. This helps smoothen the pixels inside the chromosomes and focus only on the outermost contour of the chromosomes and not the bands found inside.

	The Otsu thresholding [14] algorithm is used on the denoised grayscale image for binarization. This way, the background is separated from the foreground (chromosomes). It is important to implement adaptive thresholding to find the optimal value as this allows the algorithm to work on metaphase images of varying intensities and not just conform to a single threshold value. The next step involves the detection of chromosome outlines. One of the approaches for this is to use the Canny Edge detection [13] method. Upon using the Canny Edge algorithm on the metaphase image with the aperture size set to a value equal to 5, the boundaries of the chromosomes are clearly identified. Even finer details like the banding patterns inside the chromosome are visible to a great extent (Fig. 2). 
\begin{figure}[h]
\centering
\includegraphics[height = 4cm]{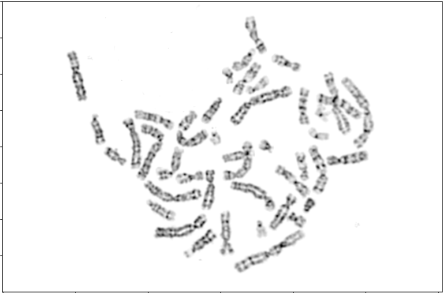}
\caption{Metaphase Image}
\includegraphics[height = 4cm]{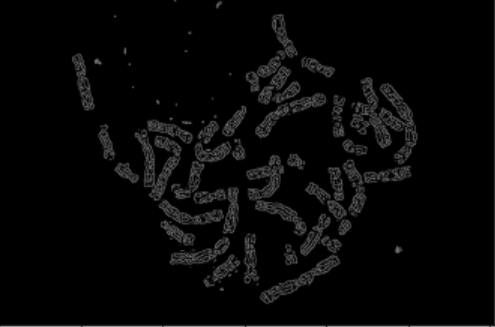}
\caption{After applying Canny Edge Detection}
\end{figure}

	As the initial step is to detect and segment the individual chromosomes, the Morphological Gradient [15] is used to detect the external boundaries alone.  
	The outlines of the chromosomes are found by applying the Morphological Gradient operator using a 2x2 kernel in the image. Morphological Gradient is essentially the difference of the dilated and eroded versions of the image which produces the outline of the boundaries (Fig. 3). Contours are found on this image and by using it as a mask, we iterate through each contour to get the chromosome Region Of Interest (ROI). Once the ROI is obtained, the chromosome is placed on a three-channel white background image. This ensures that any noise, or parts of other chromosomes are absent in that particular image. The minimum area rectangle bounding box is calculated for that chromosome, and this way multiple single chromosome images are saved from the metaphase for each contour. At this point, individual chromosomes are fully segmented and preprocessed, and clusters and overlaps have completed the initial stage of object detection. 

\begin{figure}[h]
\centering
\includegraphics[height = 4cm]{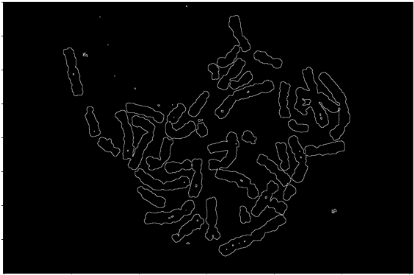}
\caption{After applying Morphological Gradient}
\end{figure}

\subsection{Detection of Overlaps and Clusters}

The next stage of the chromosome segmentation pipeline involves the detection of overlaps and clusters from the set of chromosome images that is received through the initial stage of segmentation. To achieve this, each sub-image is skeletonized after some noise reduction (Fig. 4). The skeleton of an image is its eroded, thinned down representation. The chromosomes are now shown as lines. If there exists an overlap, the skeleton lines intersect at a point. This strategy is used to detect the chromosome overlaps. The same works in case of touching and cluster chromosomes as well. 

	The presence of an intersection is found by locating a white pixel and obtaining its neighboring pixels. The neighboring pixels have a value of 1 if it's white or 0 if it's black. A list of such pixels surrounding the pixel of interest is obtained and compared with a predefined list of all possible pixel value combinations where an intersection exists. This algorithm returns the intersection coordinates if and when detected. 
	
\begin{figure}[h]
\centering
\begin{tabular}{ll}
\includegraphics[scale=1]{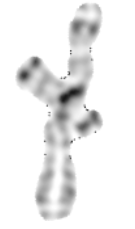}
&
\includegraphics[scale=0.4]{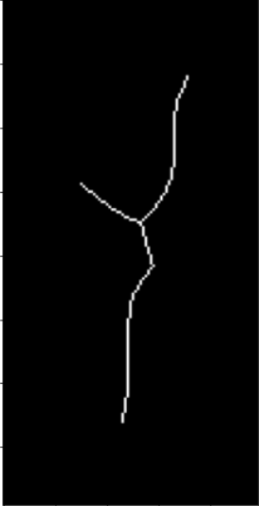}
\end{tabular}
\caption{Overlap and its skeleton}
\end{figure}

\subsection{Segmentation - Separation of Overlapping Chromosomes}

Our research has lead us to discover two types of approaches to solve this using Image Processing based techniques-
\begin{enumerate}
    \item Watershed Algorithm followed by Autoencoding to fill the missing intersection region that was initially covered by the overlap.
    \item Watershed Algorithm that considers the overlap intersection as a part of both chromosomes after separation.
\end{enumerate}
\subsubsection{Method 1}

This method works best with Q-Band chromosome images. Assume an image having two overlapping chromosomes. The watershed algorithm is used to segment the two chromosomes. Here, the markers for specifying seeds to the segmentation algorithm are provided manually.  This ensures that the user has more control on how they perceive the particular chromosome overlap. A clinician would have to decide if the intersection is due to two chromosomes crossing each other, or due to two bent chromosomes touching at the bending point. 

Once the seeds are placed appropriately at the regions on the image (Fig. 5 - left), the segments are generated accordingly, and drawn as a new image with the segments shown correspondingly (Fig. 5 - right). The regions are differentiated by changing the segment’s color. All segments belonging to a particular chromosome have the same color.

\begin{figure}[h]
\centering
\begin{tabular}{ll}
\includegraphics[scale=1]{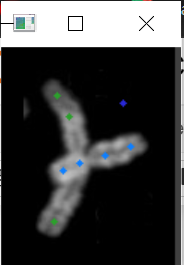}
&
\includegraphics[scale=1]{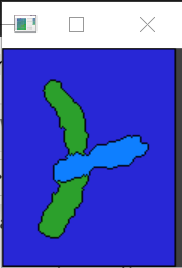}
\end{tabular}
\caption{Overlap image (left) and its segmented output (right)}
\end{figure}

Each segment is then placed on a new background image unique to each chromosome. The chromosome lying above includes the intersection region after segmentation, whereas the chromosome lying beneath has an empty region where the intersection initially was. To help fill this gap, we trained an autoencoder with the BioImLab chromosome image dataset.  An image autoencoder is a neural network that learns to first decompose the data into smaller objects and then reconstruct the original image using it to match the original as closely as possible. Autoencoders are widely used to reconstruct missing data, smoothen images, or even reduce noise. We trained the autoencoder model using the Tensorflow framework and used it to fill the gaps after the watershed segmentation if any. It was also found that autoencoding the entire image resulted in loss of valuable information like the chromosome banding pattern which is extremely important for CNN models to classify. Hence, only the missing region of the chromosome is autoencoded (Fig. 6).

\begin{figure}[h]
\centering
\begin{tabular}{ll}
\includegraphics[scale=3]{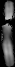}
&
\includegraphics[scale=0.65]{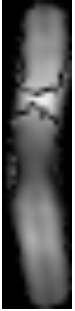}
\end{tabular}
\caption{Chromosome with a missing region (left) after autoencoding (right)}
\end{figure}

This method involves a few drawbacks. Firstly, we were able to train an accurate image autoencoder model only for the Q-Band chromosome dataset. The autoencoder for the G-Band chromosomes did not perform with a good accuracy. Secondly, the autoencoded image did not get classified correctly by the Deep CNN model at all times.

To resolve the above drawbacks we propose Method 2.
\subsubsection{Method 2}
This method supports G-Band and Q-Band chromosomes. The Watershed Algorithm described in Method 1 is implemented here as well. The seeding of chromosome segments is done in such a way that the intersection region of the overlap is specified as a separate segment from the rest of the chromosome segments (Fig. 7 - right). Finally while integrating the chromosome segments, the intersection region is integrated in both of the final separated chromosome images (Fig. 8). 

\begin{figure}[h]
\centering
\begin{tabular}{ll}
\includegraphics[scale=1]{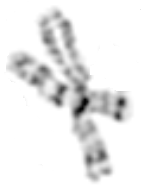}
&
\includegraphics[scale=0.6]{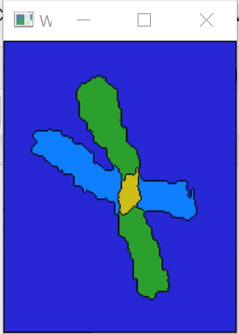}
\end{tabular}
\caption{Overlap (left) segmented with the intersection seeded separately (right)}
\end{figure}

\begin{figure}[h]
\centering
\begin{tabular}{ll}
\includegraphics[scale=0.4]{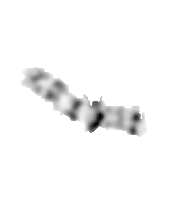}
&
\includegraphics[scale=0.4]{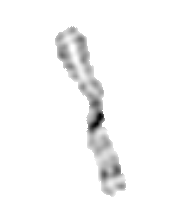}
\end{tabular}
\caption{Chromosomes after separation from the overlap}
\end{figure}

\subsection{Classification}

After completion of segmentation of chromosomes from the initial metaphase image, we receive 46 individual chromosome images (assuming a case with no abnormalities). The next step is to correctly classify the chromosomes into their respective pair numbers - there are a total of 23 pairs, with the last pair including the sex chromosomes which is X and X in case of females or X and Y in case of males.
	To do the classification, we trained a Deep Convolutional Neural Network model - SqueezeNet [11]. The SqueezeNet architecture is designed to reduce the number of parameters by "squeezing" parameters with the help of fire modules that use 1x1 convolutions. This results in a 50x reduction in model size when compared to AlexNet [12], while maintaining a good prediction accuracy. 

	The pair number for every segmented chromosome image is then predicted by using the trained CNN model, with a good accuracy. 

\subsection{Distribution Algorithm}

To further improve the final classification accuracy, the distribution algorithm is used. The idea is to take advantage of the constraint on a number of chromosomes for a given metaphase and redistribute erring predictions to best classify the chromosomes.\\
Steps-
\begin{enumerate}
    \item Get individual predictions for each pair.
    \item Find pairs that are lacking (less than 2 chromosomes are predicted for it).
    \item For each of the lacking pairs, search crowded pairs (more than 2 chromosomes are predicted for it) for an image that has the highest score for the class that the lacking pair is in.
    \item Reassign that image to the lacking pair.
    \item Repeat until there are no lacking pairs left.
\end{enumerate}

\section{Results}
The segmentation process which involves detection and separation of purely individual chromosomes from the metaphase image works efficiently and produces approximately 95\% accuracy.

In cases of segmentation involving overlap detection and separation of chromosomes, our algorithm works with most images and produces around 94\% accuracy after performing tests on multiple image segments.

The SqueezeNet model trained on all 23 pairs of chromosome images produced an accuracy of 95.75\%. The model was prepared by using an early stopping callback to prevent overfitting of the data. It trained for 119 epochs using the K-fold cross-validation strategy (5 folds). Refer Fig. 9 and Fig. 10 for visualization of the metrics. Using the distribution algorithm on the model predictions pushed the accuracy up to 98\% when the segmented images from the metaphase are given as the input to the SqueezeNet model for classification. This end-to-end segmentation-classification pipeline thus validates our approaches towards automating the karyotyping process. 

\begin{figure}[h]
\centering
\includegraphics[height = 4cm]{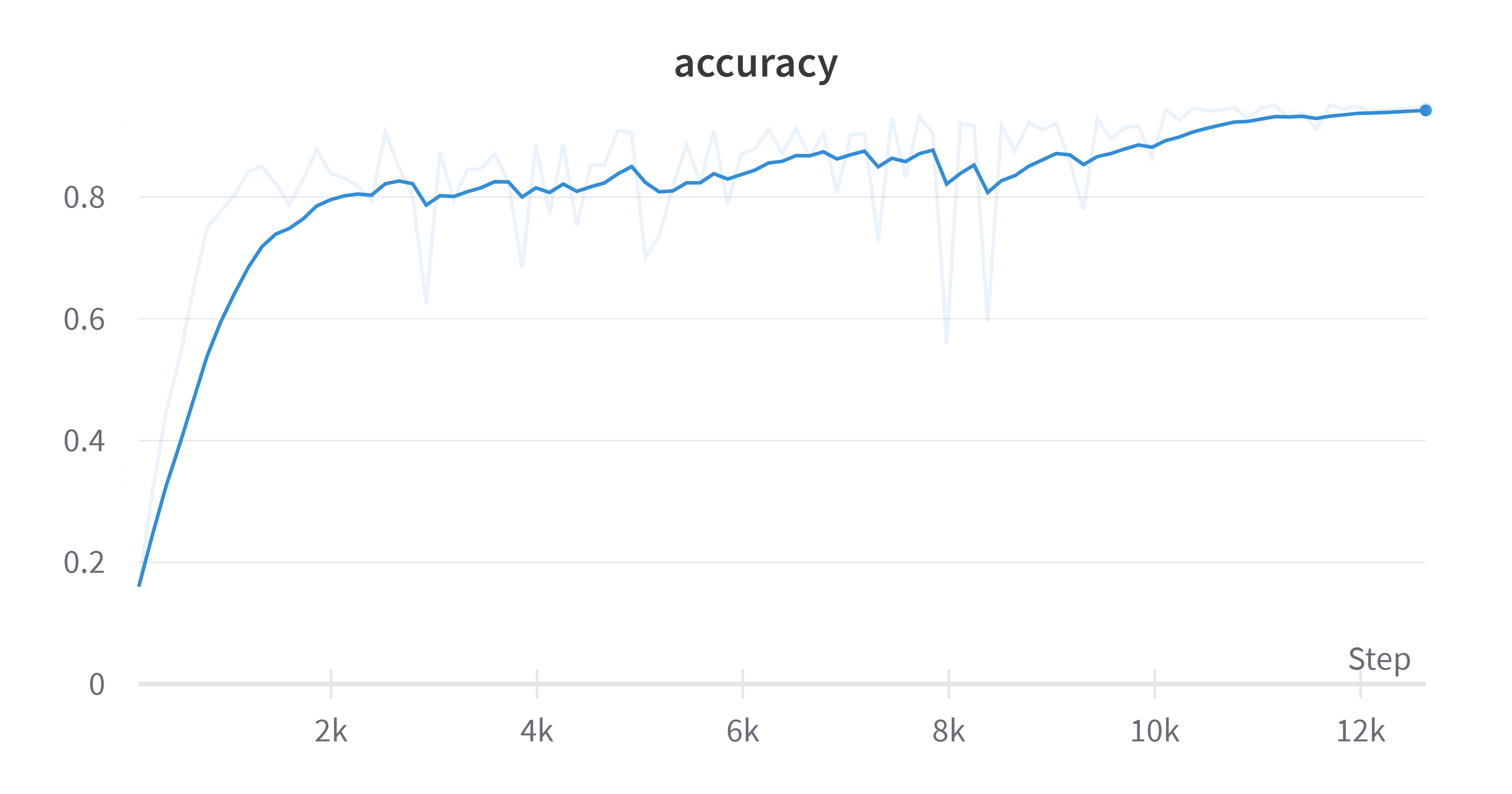}
\caption{Validation Accuracy of 95.75\% - SqueezeNet model}
\includegraphics[height = 6cm]{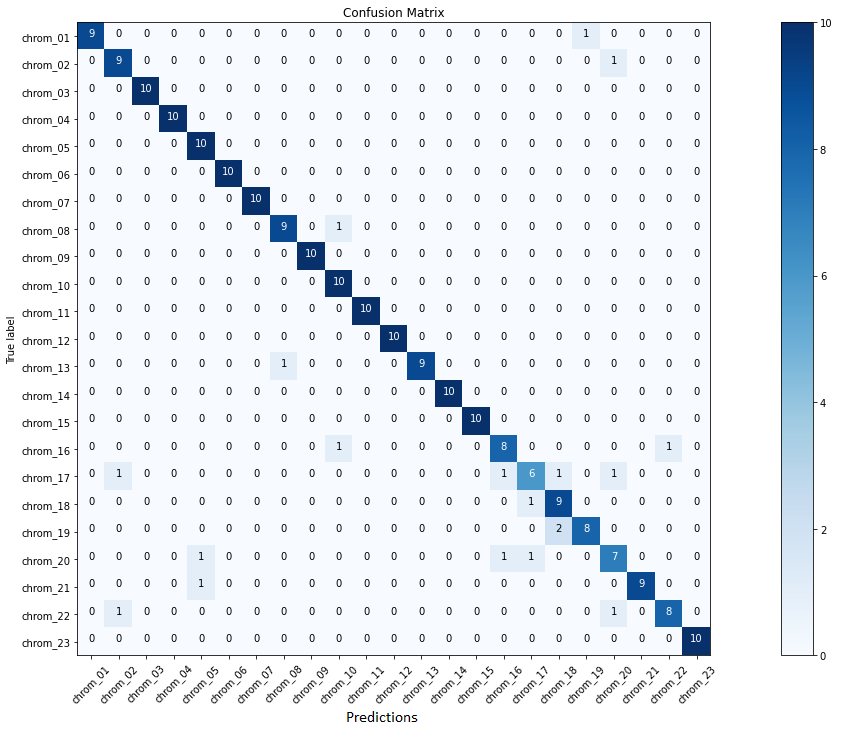}
\caption{Confusion matrix for the test data predictions}
\end{figure}

\section{Conclusion}

Karyotyping process is predominantly manual effort driven task today and is generally very tedious. The main aim of our research was to simplify segmentation of chromosomes in metaphase images, so that in particular complexities involved in the separation of touching, overlapping, and cluster chromosomes are resolved through combination of automation and very minimal manual efforts. This helps as an aid and augmentation to cytogeneticists in their task and help freshers in the field to get trained and learn the domain better and faster. We found efficient methods to initially segment the individual chromosomes, and plan to use the Canny Edge detection algorithm to further detect internal banding patterns found in chromosomes as a part of our future work.

We proposed a method that performs the overlap separation by using the watershed algorithm and segments the chromosome regions by manual seeding followed by techniques like contouring and image cropping. One of the approaches also involved the use of image autoencoding. Further we validated the chromosome class prediction accuracy after segmenting individual chromosomes from the metaphase image. This was done by training a Deep CNN - SqueezeNet model that produced and accuracy of 95.75\% on the dataset. Further through employing our distribution algorithm we could enhance the accuracy to 98\%.

\end{document}